\documentclass{article}

\usepackage{PRIMEarxiv}

\usepackage[utf8]{inputenc} 
\usepackage[T1]{fontenc}    
\usepackage{hyperref}       
\usepackage{url}            
\usepackage{booktabs}       
\usepackage{amsfonts}       
\usepackage{nicefrac}       
\usepackage{microtype}      
\usepackage{lipsum}
\usepackage{fancyhdr}       
\usepackage{graphicx}       
\graphicspath{{media/}}     
\usepackage{adjustbox}
\usepackage{amsmath}
\title{
Innovative LSGTime Model for Crime Spatiotemporal Prediction Based on MindSpore Framework 
}

\author{
  Zhenkai Qin$^{1,2,3}$ \thanks{These authors contributed equally to this work.} \\
  $^{1}$School of Computing and Information \\
  $^{2}$Network Security Research Center \\
  $^{3}$Big Data and Policing Technology Laboratory\\
  Guangxi Police College\\
  Nanning, Guangxi, China \\
  \texttt{qinzhenkai@gxjcxy.edu.cn} \\
  \And
  Baozhong Wei \\
  Institute of Software \\
  Chinese Academy of Sciences \\
  Beijing , China \\
  \texttt{baozhong@isrc.iscas.ac.cn} \\
  \And
  Caifeng Gao  \\
  Institute of Software \\
  Chinese Academy of Sciences \\
  Beijing , China \\
  \texttt{caifeng@isrc.iscas.ac.cn} \\
}

\begin{document}
\maketitle

\begin{abstract}
With the acceleration of urbanization, the spatiotemporal characteristics of criminal activities have become increasingly complex. Accurate prediction of crime distribution is crucial for optimizing the allocation of police resources and preventing crime. This paper proposes LGSTime, a crime spatiotemporal prediction model that integrates Long Short-Term Memory (LSTM), Gated Recurrent Unit (GRU), and the Multi-head Sparse Self-attention mechanism. LSTM and GRU capture long-term dependencies in crime time series, such as seasonality and periodicity, through their unique gating mechanisms. The Multi-head Sparse Self-attention mechanism, on the other hand, focuses on both temporal and spatial features of criminal events simultaneously through parallel processing and sparsification techniques, significantly improving computational efficiency and prediction accuracy. The integrated model leverages the strengths of each technique to better handle complex spatiotemporal data. Experimental findings demonstrate that the model attains optimal performance across four real - world crime datasets. In comparison to the CNN model, it exhibits performance enhancements of 2.8\%, 1.9\%, and 1.4\% in the Mean Squared Error (MSE), Mean Absolute Error (MAE), and Root Mean Squared Error (RMSE) metrics respectively. These results offer a valuable reference for tackling the challenges in crime prediction. 
\end{abstract}


\section{Introduction}
Crime does not occur entirely at random but exhibits certain patterns in both time and space. With the acceleration of urbanization and the increasing complexity of social structures, the spatiotemporal characteristics of criminal activities have become more intricate. Accurate prediction of crime spatiotemporal distribution is of great significance for optimizing the allocation of police resources, preventing the occurrence of crimes, and maintaining social security. In recent years, with the development of big data technologies, a vast amount of crime data has been collected and stored, providing the possibility for utilizing advanced data mining and machine learning techniques to predict crime spatiotemporal patterns \cite{1crime_data_analysis_2021}.

Long Short-Term Memory (LSTM) and Gated Recurrent Unit (GRU) are two important variants of Recurrent Neural Networks (RNNs) that have significant advantages in processing time series data and have thus been widely applied in many predictive fields. LSTM, initially proposed by Hochreiter and Schmidhuber in 1997 \cite{2lstm_1997}, addresses the vanishing gradient problem encountered by traditional RNNs when dealing with long sequences through its gating mechanism \cite{2lstm_1997}. By the coordinated action of the input gate, forget gate, and output gate, LSTM can flexibly control the flow of information, thereby effectively capturing long-term dependencies in time series. This characteristic has enabled LSTM to achieve remarkable results in areas such as language modeling, speech recognition, and financial market prediction\cite{3lstm_finance_2020}. GRU, introduced by Cho et al. in 2014 \cite{4gru_2014} as a simplified version of LSTM, reduces the number of model parameters while maintaining effective modeling capabilities for time series data by introducing update and reset gates \cite{4gru_2014}. With a more concise structure and faster training speed, GRU has gradually become a strong alternative to LSTM in many practical applications, performing comparably or even superior to LSTM in fields such as natural language processing, time series prediction, and biomedical signal analysis \cite{5gru_vs_lstm_2019}.

In the field of crime prediction, the application of LSTM and GRU has also attracted increasing attention. The time series data of criminal events typically exhibit complex long-term dependencies, such as seasonality, periodicity of criminal behavior, and causal relationships between events. LSTM and GRU can effectively capture these long-term dependencies, thus providing powerful tools for crime spatiotemporal prediction \cite{6crime_prediction_lstm_gru_2020}. For example, by analyzing the time series of historical crime data, LSTM and GRU can predict the probability of crime occurrence in a future time period, helping law enforcement agencies to deploy police resources in advance and prevent crimes \cite{7crime_hotspots_lstm_2019}. However, despite the achievements of LSTM and GRU in the field of crime prediction, there are still some limitations when they are used separately. The complex model structures of LSTM and GRU require substantial computational resources and time for training, and they are sensitive to data noise, which can be easily affected by outliers and thus reduce the accuracy of prediction \cite{8limitations_lstm_gru_2021}. Moreover, their scalability in handling large-scale data is relatively poor, making it difficult to efficiently process large datasets. More importantly, LSTM and GRU mainly focus on the long-term dependencies in time series data and fail to fully capture the spatiotemporal dependencies of criminal events, resulting in limited prediction accuracy \cite{9spatiotemporal_crime_2022}.

To overcome the limitations of LSTM and GRU in crime prediction, the Multi-head Sparse Self-attention(MHSA) mechanism has been gradually introduced into this field in recent years. As a deep learning method based on the attention mechanism, Multi-head Sparse Self-attention can simultaneously focus on multiple features of criminal events in both time and space while improving computational efficiency through sparsification \cite{10multi_head_attention_2023}. This technique can automatically learn the importance weights of criminal events across different time scales and spatial regions, thereby more accurately capturing the spatiotemporal patterns of criminal activities. For example, it can identify high-crime patterns in specific areas during certain time periods and make predictions accordingly. The advantage of Multi-head Sparse Self-attention lies in its ability to process data in parallel, which significantly enhances computational efficiency and enables better capture of both local and global features in the data. Through sparsification, this mechanism can reduce computational load while retaining key information, thus performing well on large-scale datasets. In addition, Multi-head Sparse Self-attention is more robust to data noise and can better handle outliers, thereby improving the accuracy of prediction \cite{11robustness_attention_2023}.

To better address the aforementioned issues, this study proposes the LGSTime model, which integrates LSTM, GRU, and Multi-head Sparse Self-attention mechanisms to leverage their respective strengths and construct a more powerful and accurate crime spatiotemporal prediction model. This hybrid model can not only effectively utilize the long-term dependencies in time series data but also better handle the complex spatial distribution and interrelations of criminal events. While LSTM and GRU capture long-term dependencies in the temporal dimension, Multi-head Sparse Self-attention focuses on multiple features in both time and space, thereby providing a more comprehensive understanding of the spatiotemporal patterns of criminal events. The main contributions of this paper are as follows:
\begin{itemize}

\item This study proposes an integrated crime spatiotemporal prediction model combining LSTM, GRU, and Multi-head Sparse Self-attention mechanisms. This integrated model fully exploits the advantages of each technique, thereby enhancing the model's ability to capture complex spatiotemporal dependencies in crime data.

\item The introduction of the Multi-head Sparse Self-attention mechanism significantly improves the model's computational efficiency and prediction accuracy. This mechanism enables the model to focus on the most relevant features in both time and space, thereby providing more precise crime predictions.

\item The proposed model is rigorously validated using four real-world crime datasets. The results demonstrate that the model achieves high accuracy in predicting crime spatiotemporal patterns.

\end{itemize}

\section{Related Work}
\subsection{LSTM In Crime Prediction}
As a classic variant of Recurrent Neural Networks (RNNs), Long Short-Term Memory (LSTM) networks have been widely applied in the field of crime prediction, thanks to their unique gating mechanisms that effectively address the vanishing gradient problem encountered by traditional RNNs when processing long sequences. Through the coordinated action of the input gate, forget gate, and output gate, LSTM can flexibly control the flow of information, thereby capturing long-term dependencies in time series data. This capability enables LSTM to analyze temporal features of criminal events, such as seasonality, periodicity, and causal relationships between events, thus providing a powerful tool for crime prediction \cite{12crime_prediction_lstm_gru_2020}.

In terms of technical implementation, LSTM models typically take the time series data of criminal events as input and model the temporal features of historical crime data to predict the probability of crime occurrence in future time periods. For example, Smith et al. (2018) proposed an LSTM-based time series prediction model for theft crimes. By analyzing the temporal features of historical crime data, this model accurately captured the long-term dependencies of criminal events in the temporal dimension, thereby more accurately predicting future crime trends \cite{13crime_hotspots_lstm_2019}. In addition, many researchers have further enhanced the predictive performance of LSTM by combining it with other techniques, such as Convolutional Neural Networks (CNNs) and attention mechanisms \cite{14lstm_attention_2021}.

However, despite the achievements of LSTM in crime prediction, there are still some limitations when it is used alone. First, the complex structure of LSTM requires substantial computational resources and time for training, making it inefficient for handling large-scale datasets \cite{15limitations_lstm_2022}. Second, LSTM primarily focuses on long-term dependencies in time series data and struggles to fully capture spatiotemporal dependencies of criminal events, resulting in limited prediction accuracy \cite{16spatiotemporal_crime_2022}. Moreover, LSTM is sensitive to data noise and can be easily affected by outliers, thereby reducing the accuracy of predictions \cite{17robustness_lstm_2023}.

\subsection{GRU In Crime Prediction}
The Gated Recurrent Unit (GRU), as an efficient variant of Recurrent Neural Networks (RNNs), has been widely applied in the field of crime prediction due to its concise structure and high computational efficiency. By introducing update and reset gates, the GRU effectively addresses the vanishing gradient problem encountered by traditional RNNs when processing long sequences \cite{18gru_crime_review_2020}. Its working principle involves two key steps: First, the reset gate determines the candidate hidden state, which combines the current input and the hidden state from the previous time step. Second, the update gate integrates the candidate hidden state with the previous hidden state to generate the current hidden state. This mechanism enables the GRU to flexibly retain or discard past information, thereby better capturing long-term dependencies in time series data \cite{19gru_long_term_2021}.

In crime prediction, the GRU has been extensively used to model the temporal features of criminal events. For instance, some studies have proposed GRU-based crime prediction models that analyze the time series of historical crime data to accurately capture the seasonality and periodicity of criminal events, thereby predicting the probability of future crimes \cite{20gru_crime_models_2021}. Moreover, the GRU has been combined with other techniques, such as graph autoencoders and convolutional neural networks, to further enhance the predictive performance of the models. For example, the Graph Autoencoder and GRU Neural Network (GAERNN) model has shown remarkable performance in handling complex spatiotemporal data, enabling more accurate predictions of crime distributions in both space and time \cite{21gaernn_2022}.

However, the application of GRU in crime prediction also has some limitations. First, despite its relatively simple structure, the GRU is sensitive to the choice of hyperparameters, which require careful tuning to achieve optimal performance \cite{22gru_hyperparameters_2022}. Second, when processing very long sequences, the GRU may still encounter vanishing gradient problems, affecting its ability to model complex sequence dependencies \cite{23gru_vanishing_gradient_2023}. Additionally, the GRU has relatively low interpretability, with its internal information flow and gating mechanisms being difficult to understand, which to some extent limits its explainability in practical applications. Finally, as a variant of RNNs, the GRU still cannot perform parallel computation, which poses significant challenges when dealing with large-scale data \cite{24gru_large_scale_2023}.

\subsection{MHSA In Crime Prediction}

Multi-head Sparse Self-attention (MHSA) is an enhanced method based on the Transformer architecture, specifically designed for efficient processing of large-scale sequential data. Its core innovation lies in the integration of multi-head attention mechanisms with sparsity techniques, which significantly reduces computational complexity and memory usage while maintaining the model's powerful expressive capabilities \cite{25mhsa_transformer_2021}.

In crime prediction, this mechanism processes crime data time series in parallel through multiple attention heads, with each head focusing on different parts of the sequence. This enables the simultaneous capture of long-range dependencies and local features of crime events \cite{26mhsa_features_2022}. For instance, when analyzing urban crime data, MHSA can concurrently monitor crime patterns across different regions and their temporal variations, thereby more accurately identifying crime hotspots and trends \cite{27mhsa_crime_hotspots_2023}. Additionally, the sparsity technique allows the model to skip unnecessary computations when calculating attention weights, further enhancing efficiency \cite{28mhsa_sparsity_2023}.

Despite its impressive performance in crime prediction, MHSA has certain limitations. First, its sparsity strategy may lead to the omission of some critical information, particularly when dealing with complex spatiotemporal data, potentially losing key local features \cite{29mhsa_limitations_2023}. Second, the multi-head attention mechanism inherently increases model complexity, resulting in a larger number of parameters and greater difficulty in training and tuning \cite{30mhsa_complexity_2023}. Moreover, although sparsity reduces computational complexity, memory usage remains high when processing extremely long sequences \cite{31mhsa_memory_2023}. Finally, the interpretability of this mechanism is limited, making it challenging to intuitively understand the specific functions of each attention head, which somewhat restricts its practical applicability in real-world scenarios \cite{32mhsa_interpretability_2023}.
\section{Methodology}
\begin{figure}[h]
  \centering
  \includegraphics[width=\textwidth]{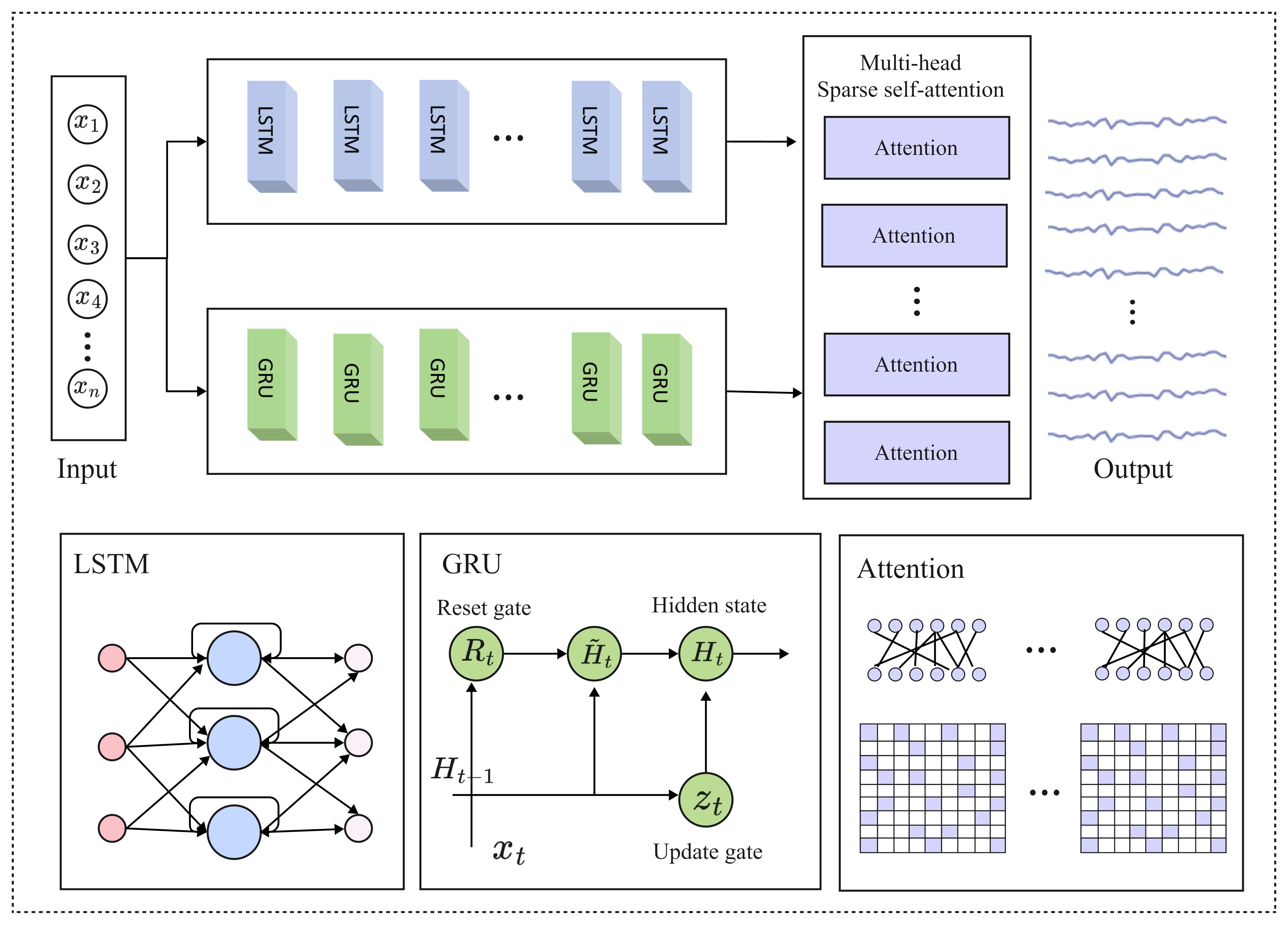}
  \caption{The LGSTime model consists of three modules: LSTM, capable of capturing long-term dependencies in time series data; GRU, focused on short-term dependencies; and a Multi-head Sparse self-attention module that identifies complex relationships between different features within the input sequence.}
  \label{liucheng.png}
\end{figure}

As illustrated in Figure \ref{liucheng.png}, the LGSTime model is composed of three key modules: LSTM, GRU, and the Multi-head Sparse Attention module. The LSTM module captures long-term dependencies in time series data through its gating mechanisms, providing the model with robust long-term predictive capabilities. The GRU module focuses on short-term dependencies, rapidly adapting to immediate changes in criminal events to enhance the model's short-term forecasting agility. The Multi-head Sparse Attention module identifies complex relationships between different features in the input sequence, improving prediction accuracy and robustness by processing multiple attention heads in parallel. The integration of these three modules enables the LGSTime model to excel in crime prediction, handling both long-term trends and short-term fluctuations, as well as the interplay between multiple features.
\subsection{LSTM Gate Mechanism}
The LSTM preserves long-term memory through cell state $C_t$, with core formulas:
\begin{equation}
f_t = \sigma(W_f[h_{t-1},x_t] + b_f)\\   
\end{equation}
In this equation, $f_t$ represents the activation vector of the forget gate at time step $t$. The sigmoid activation function is denoted by $\sigma$. $W_f$ is the weight matrix associated with the forget gate, $h_{t - 1}$ stands for the hidden state at the previous time step, $x_t$ is the input at time step $t$, and $b_f$ is the bias term of the forget gate.

\begin{equation}
i_t = \sigma(W_i[h_{t-1},x_t] + b_i)\\    
\end{equation}
 Where,
$ i_t $ is the input gate activation vector at time step $ t $;
 $ \sigma $ denotes the sigmoid activation function;
 $ W_i $ is the weight matrix for the input gate;
 $ h_{t-1} $ is the hidden state at the previous time step;
 $ x_t $ is the input at time step $ t $;
 $ b_i $ is the bias term for the input gate.
\begin{equation}
\tilde{C}t = \tanh(W_c[h{t-1},x_t] + b_c)\\  
\end{equation}
Where,
 $ \tilde{C_t} $ is the candidate cell state at time step $ t $;
 $ \tanh $ denotes the hyperbolic tangent activation function;
 $ W_c $ is the weight matrix for the candidate cell state;
 $ h_{t-1} $ is the hidden state at the previous time step;
 $ x_t $ is the input at time step $ t $;
 $ b_c $ is the bias term for the candidate cell state.
\begin{equation}
 C_t = f_t \odot C_{t-1} + i_t \odot \tilde{C}t \\  
\end{equation}
Here, $C_t$ is the updated cell state at time step $t$;
$\odot$ denotes the element-wise multiplication (Hadamard product).

\begin{equation}
 o_t = \sigma(W_o[h{t-1},x_t] + b_o)\\ 
\end{equation}
Where,
 $ o_t $ is the output gate activation vector at time step $ t $;
$ \sigma $ denotes the sigmoid activation function;
 $ W_o $ is the weight matrix for the output gate;
 $ h_{t-1} $ is the hidden state at the previous time step;
 $ x_t $ is the input at time step $ t $;
 $ b_o $ is the bias term for the output gate.
\begin{equation}
h_t = o_t \odot \tanh(C_t)\\
\end{equation}
Here, $ h_t $ represents the hidden state at time step $ t $; $ \odot $ signifies the element-wise multiplication, also known as the Hadamard product; $ \tanh $ stands for the hyperbolic tangent activation function.

 \subsection{GRU Gate Mechanism}

The Gated Recurrent Unit (GRU) achieves efficient computation by merging gate units. The mathematical formulations for the GRU are as follows:

\begin{equation}
z_t = \sigma(W_z [h_{t-1}, x_t] + b_z)  \\
\end{equation}
At time step \( t \), the update gate \( z_t \) controls how much past information is passed along. Here, \( \sigma \) denotes the sigmoid activation function, \( W_z \) is the weight matrix for the update gate, \( h_{t-1} \) represents the hidden state from the previous time step, \( x_t \) is the input at the current time step, and \( b_z \) is the bias term for the update gate.
\begin{equation}
r_t = \sigma(W_r [h_{t-1}, x_t] + b_r)  \\
\end{equation}
The reset gate $r_t$ at time step $t$ determines how much of the past information should be forgotten, where $W_r$ is the weight matrix for the reset gate and $b_r$ is the corresponding bias term.
\begin{equation}
\tilde{h}_t = \tanh(W_h [r_t \odot h_{t-1}, x_t] + b_h)  \\
\end{equation}
At time step \(t\), the candidate hidden state \(\tilde{h}_t\) is obtained by applying the hyperbolic tangent function to the weighted sum of the reset previous hidden state (\(r_t \odot h_{t - 1}\)) and the current input (\(x_t\)). Here, \(W_h\) serves as the weight matrix, \(b_h\) is the bias term, and \(\odot\) represents element - wise multiplication (Hadamard product), thus integrating the transformed past information with the current input. 
\begin{equation}
h_t = (1 - z_t) \odot h_{t-1} + z_t \odot \tilde{h}_t \\
\end{equation}
The final hidden state $h_t$ at time step $t$ is computed as a weighted sum of the previous hidden state and the candidate hidden state, determined by the update gate, where $h_t$ retains some of the past information ($(1 - z_t) \odot h_{t-1}$) and adds new information ($z_t \odot \tilde{h}_t$).
\subsection{Multi-head Sparse Self-attention}
In the multi-head Sparse self-attention mechanism, the input sequence \( X \) is subjected to three distinct linear transformations to produce the Query, Key, and Value matrices, denoted as \( Q \), \( K \), and \( V \) respectively. For each individual attention head, the input sequence \( X \) is transformed via linear operations using separate weight matrices \( W^Q \), \( W^K \), and \( W^V \), which generate the corresponding \( Q \), \( K \), and \( V \) matrices. The mathematical formulations are as follows:

\begin{equation}
Q = XW^Q, \quad K = XW^K, \quad V = XW^V
\end{equation}

Here, \(W^Q\), \(W^K\), and \(W^V\) represent projection matrices, each with dimensions of \(d_{model} \times d_{model}\). The dimension \(d_{model}\) refers to the model dimension, which is also the feature dimension of the input sequence \(X\). Subsequently, the similarity \(S_i\) is calculated as follows:
\begin{equation}
S_i = \frac{Q_i K_i^T}{\sqrt{d_h}} 
\end{equation}

Where \(d_h\) is the feature dimension of each head, we define the local attention masking matrix \(M \in \{0,1\}^{n \times n}\).
\begin{equation}
M_{j,k} = 
\begin{cases} 
1 & \text{if } |j - k| \leq k \\
0 & \text{otherwise}
\end{cases}
\end{equation}

Here, \( k \) represents the \texttt{sparse\_factor} parameter, which governs the window radius. The formula for the masked attention weights is as follows:

\begin{equation}
\tilde{S}_i = S_i \odot M + (-\infty)(1 - M)
\end{equation}

\begin{equation}
A_i = \text{Softmax}(\tilde{S}_i)
\end{equation}

Where \( S_i \) represents the original attention scores, and \( M \) denotes the mask matrix, with masked positions set to 0 and unmasked positions set to 1, the formula for the sparse attention mechanism is as follows:
\begin{equation}
\text{head}_i = A_i V_i
\end{equation}

\begin{equation}
\text{Output} = \text{Concat}(\text{head}_1, \ldots, \text{head}_h) W^O
\end{equation}

Where \( W^O \in \mathbb{R}^{d_{\text{model}} \times d_{\text{model}}} \) denotes the output projection matrix.

\section{Experiments}
In this study, we utilize Mean Squared Error (MSE), Mean Absolute Error (MAE), and Root Mean Squared Error (RMSE) as three essential criteria to assess the predictive accuracy of the LGSTime model. The respective formulas for computing these metrics are detailed below:
\begin{equation}
\text{MSE} = \frac{1}{n} \sum_{i=1}^{n} \left( y_{i} - \hat{y}_{i} \right)^2
\end{equation}

\begin{equation}
\text{MAE} = \frac{1}{n} \sum_{i=1}^{n} \left| y_{i} - \hat{y}_{i} \right|
\end{equation}

\begin{equation}
\text{RMSE} = \sqrt{\frac{1}{n} \sum_{i=1}^{n} (y_i - \hat{y}_i)^2}
\end{equation}
Here, \( y_{i} \) represents the actual observed value, \( \hat{y}_{i} \) is the predicted value of the model, and \( n \) is the total number of data points. The lower the values of Mean Squared Error (MSE) and Mean Absolute Error (MAE), the closer the predicted values are to the actual values. Moreover, a lower Root Mean Squared Error (RMSE) indicates a better fit of the model to the data.

\subsection{Datasets}
According to data from \url{https://www.neighborhoodscout.com/ca/los-angeles/crime}, Los Angeles has a crime rate significantly higher than the national average in the United States. In terms of safety, Los Angeles is safer than only 6\% of U.S. cities. To scientifically evaluate the performance of LGSTime in spatiotemporal crime sequence prediction, this study collected and utilized historical crime data from Los Angeles between January 1, 2020, and September 10, 2023. During the data processing phase, we initially excluded some non-essential indicators to focus on core data, reduce interference, and enhance data processing efficiency. Subsequently, the data was divided into four segments by year, corresponding to the crime data for 2020, 2021, 2022, and 2023. Finally, 2,500 records were randomly selected from each segment to form four datasets, named DS1, DS2, DS3, and DS4. To ensure the scientific validity and effectiveness of model training, the entire dataset was systematically divided into training, validation, and testing sets in chronological order, with a ratio of 7:1:2. The data used in this study is rich in dimensions, covering 12 key indicators. Additionally, the data includes various common types of crimes, such as theft, assault, and motor vehicle theft. This data provides a solid foundation for accurately assessing the performance of LGSTime.
\subsection{Implementation Details}
In the training process of this study, we employed the Mean Squared Error (MSE) loss function and utilized the ADAM optimizer for model training, with the learning rate set to 0.00001. To enhance the model’s generalization capability, we adopted a weight decay of 0.1 to prevent overfitting. Training was terminated after reaching 100 epochs. To ensure the robustness of the experimental results, we conducted each experiment three times and executed them using the Mindspore framework on an NVIDIA Tesla V100 32GB GPU.
\subsection{Multivariate Result}
In the experimental settings of multivariate analysis (based on the Mindspore framework), we conducted experiments on four distinct datasets. The results demonstrate that the LGSTime model consistently achieved state-of-the-art performance across the majority of baseline and prediction horizon configurations (see Table \ref{duobianlian}). Specifically, under the configuration with an input length of 96 and a prediction length of 1, the LGSTime model achieved significant performance improvements compared to previous state-of-the-art results across all datasets. On the DS1 dataset, the LGSTime model reduced the mean squared error (MSE) by 0.19\% (from 0.1001 to 0.999); on DS2, the MSE was reduced by 0.80\% (from 0.991 to 0.983); on DS3, the MSE was reduced by 0.21\% (from 0.939 to 0.937); and on DS4, the MSE was reduced by 1.6\% (from 1.060 to 1.043). Furthermore, by calculating the number of times the LGSTime model achieved the best performance across three metrics—MSE, MAE, and RMSE—the model demonstrated the best performance 4 times in MSE, 2 times in MAE, and 4 times in RMSE. These results provide compelling evidence of the LGSTime model's superiority in the task of crime forecasting, showcasing its significant performance advantages across multiple key metrics and highlighting its potential to provide robust support for research and applications in this domain.
 \begin{table}[h]
 \centering
 \begin{adjustbox}{max width=1\textwidth}
 \begin{tabular}{c c cc cc cc cc cc cc cc cc cc cc cc}
 \toprule
 \multicolumn{2}{c}{\textbf{Model}}  & \multicolumn{3}{c}{\textbf{LGSTime}} & \multicolumn{3}{c}{\textbf{CNN}} & \multicolumn{3}{c}{\textbf{RNN}} & \multicolumn{3}{c}{\textbf{GRU}} \\
 \multicolumn{2}{c}{\textbf{Metric}}  & \textbf{MSE} & \textbf{MAE} & \textbf{RMSE} & \textbf{MSE} & \textbf{MAE} & \textbf{RMSE} & \textbf{MSE} & \textbf{MAE} & \textbf{RMSE} & \textbf{MSE} & \textbf{MAE} & \textbf{RMSE} \\
 \midrule
 \multicolumn{2}{c}{\textbf{DS1}} & \textbf{0.999} & \underline{0.776} & \textbf{1.000} & 1.061 & 0.805 & 1.030 & 1.002 & \textbf{0.774} & \underline{1.001} & \underline{1.001} & 0.776 & \textbf{1.000} \\
 \midrule
 \multicolumn{2}{c}{\textbf{DS2}} & \textbf{0.983} & \textbf{0.759} & \textbf{0.992} & 0.992 & 0.761 & 0.996 & \underline{0.991} & \underline{0.760} & \underline{0.995} & 0.992 & 0.761& 0.996 \\
 \midrule
 \multicolumn{2}{c}{\textbf{DS3}} & \textbf{0.937} & \textbf{0.750} & \textbf{0.968} & 0.959 & 0.770 & 0.979 & 0.937 & \underline{0.751} & \textbf{0.968}  & \underline{0.939} & 0.753 & \underline{0.969} \\
 \midrule
 \multicolumn{2}{c}{\textbf{DS4}} & \textbf{1.043} & \textbf{0.799} & \textbf{1.021} & 1.067 & 0.808 & 1.033 & \underline{1.060} & \underline{0.804} & \underline{1.029} & 1.064 & 0.805 & 1.032 \\
 \midrule
 \multicolumn{2}{c}{\textbf{Count}} & 4 & 2 & 4 & 0 & 0 & 0 & 0 & 2 & 1 & 0 & 0 & 1 \\
 \bottomrule
 \end{tabular}
 \end{adjustbox}
 \caption{Multivariate results for four different datasets with an input length of 96 and a prediction length of 1. The MSE Reduction refers to the percentage decrease in mean squared error (MSE) of LGSTime compared to other models. The best average results are highlighted in \textbf{bold}, while the second-best results are indicated with an \underline{underline.}}
 \label{duobianlian}
 \end{table}

\subsection{Ablation Research}
This study evaluated the impact of the Multi-Head Sparse Attention (MHSA) module on model performance through ablation experiments (based on the Mindspore framework). Two variants of LGSTime were tested: one with LSTM+GRU, where the MHSA mechanism was completely removed; and another using only the LSTM model for prediction. The experiments were conducted on four datasets, with the results presented in Table \ref{xiaorong}. LGSTime achieved the best performance in 11 out of 12 test cases. These results confirm the superior performance of the MHSA module in the task of crime prediction. The superiority of the MHSA module can be attributed to its unique design and functionality. First, the multi-head mechanism enables the model to model these factors from different dimensions, thereby gaining a more comprehensive understanding of the complex structures within the data. Second, the sparse attention mechanism effectively reduces computational complexity while maintaining focus on key information. Additionally, the MHSA module can dynamically adjust attention weights, automatically assigning more weight to more important features based on the characteristics and contextual information of the input data, thereby enhancing the model's adaptability and accuracy.
\begin{table}[h]
\centering
\begin{adjustbox}{max width=1\textwidth}
\begin{tabular}{c c cc cc cc cc cc cc cc cc cc cc cc}
\toprule
\multicolumn{2}{c}{\textbf{Model}}  & \multicolumn{3}{c}{\textbf{LGSTime}} & \multicolumn{3}{c}{\textbf{LSTM+GRU}} & \multicolumn{3}{c}{\textbf{LSTM}} \\
\multicolumn{2}{c}{\textbf{Metric}}  & \textbf{MSE} & \textbf{MAE} & \textbf{RMSE} & \textbf{MSE} & \textbf{MAE} & \textbf{RMSE} & \textbf{MSE} & \textbf{MAE} & \textbf{RMSE} \\
\midrule
\multicolumn{2}{c}{\textbf{DS1}} & \textbf{0.999} & \underline{0.776} & \textbf{1.000} & 1.005 & 0.778 & 1.002 & \underline{1.001} & \textbf{0.775} & \underline{1.001} \\
\midrule
\multicolumn{2}{c}{\textbf{DS2}} & \textbf{0.983} & \textbf{0.759} & \textbf{0.992} & 0.987 & \underline{0.760} & \underline{0.993} & \underline{0.986} & 0.761 & \underline{0.993} \\
\midrule
\multicolumn{2}{c}{\textbf{DS3}} & \textbf{0.937} & \textbf{0.750} & \textbf{0.968} & 0.947 & 0.758 & 0.973 & \underline{0.941} & \underline{0.751} & \underline{0.970} \\
\midrule
\multicolumn{2}{c}{\textbf{DS4}} & \textbf{1.043} & \textbf{0.799} & \textbf{1.021} & \underline{1.051} & \underline{0.803} & \underline{1.025} & 1.058 & 0.804 & 1.028 \\
\midrule
\multicolumn{2}{c}{\textbf{Count}} & 4 & 3 & 4 & 0 & 0 & 0 & 0 & 1 & 0 \\
\bottomrule
\end{tabular}
\end{adjustbox}
\caption{Ablation study results for four different datasets with an input length of 96 and a prediction length of 1. MSE Reduction refers to the percentage decrease in mean squared error (MSE) of LGSTime compared to other models. The best average results are shown in \textbf{bold}, and the second-best in \underline{underlined.}}
\label{xiaorong}
\end{table}

\section{Conclusions}
With the acceleration of urbanization and the increasing complexity of social structures, the spatiotemporal characteristics of criminal activities have become more intricate, making accurate prediction of crime distribution crucial for optimizing police resource allocation and crime prevention. This paper proposes the LGSTime model, which integrates LSTM, GRU, and the Multi-head Sparse Self-attention mechanism, aiming to leverage the strengths of each technology to build a more robust and accurate spatiotemporal crime prediction model. LSTM and GRU capture long-term dependencies in crime time series, such as seasonality and periodicity, through their unique gating mechanisms, while the Multi-head Sparse Self-attention mechanism simultaneously focuses on both temporal and spatial features of criminal events through parallel processing and sparsification techniques, significantly improving computational efficiency and prediction accuracy. The integrated model fully utilizes the advantages of each technology to better handle complex spatiotemporal data. Experimental results demonstrate that the model achieves optimal performance on four real-world crime datasets, providing a valuable reference for addressing crime prediction challenges. In the future, the LGSTime model is expected to make breakthroughs in areas such as multimodal data fusion, lightweight deployment, and cross-domain applications, offering broader support for public safety governance.
\section*{Acknowledgments}
Thanks for the support provided by the MindSpore Community.

\end{document}